\documentclass[11pt]{article}

\usepackage[preprint]{acl}

\usepackage{times}
\usepackage{latexsym}

\usepackage[T1]{fontenc}

\usepackage[utf8]{inputenc}

\usepackage{microtype}

\usepackage{inconsolata}

\usepackage{graphicx}

\title{Instructions for *ACL Proceedings}

\usepackage{amsmath,amsfonts,bm}

\def\eqref#1{equation~\ref{#1}}

\def\1{\bm{1}}

\DeclareMathAlphabet{\mathsfit}{\encodingdefault}{\sfdefault}{m}{sl}
\SetMathAlphabet{\mathsfit}{bold}{\encodingdefault}{\sfdefault}{bx}{n}

\usepackage{hyperref}
\usepackage{url}

\usepackage{booktabs}       
\usepackage{amsfonts}       
\usepackage{nicefrac}       
\usepackage{microtype}      
\usepackage{xcolor}         
\usepackage{threeparttable}

\usepackage{xcolor}
\usepackage{colortbl}
\usepackage{enumitem}
\usepackage{makecell}
\definecolor{colorTab}{rgb}{0.9,0.9,0.98}
\definecolor{color3}{gray}{0.95}

\definecolor{css}{rgb}{0.7529, 0, 0}
\definecolor{fss}{rgb}{0, 0.7, 0.3}
\definecolor{pbp}{rgb}{0.2, 0.2, 0.6}
\usepackage[utf8]{inputenc}
\usepackage[T1]{fontenc}
\usepackage[american]{babel}
\usepackage{microtype}

\setlist[itemize]{noitemsep, topsep=0pt, left=0pt}

\usepackage{amsmath}
\usepackage{amsthm}
\usepackage{amsfonts}
\usepackage{amssymb}
\usepackage{mathtools}
\usepackage{nicefrac}
\usepackage{algorithm}
\usepackage{multicol}
\usepackage{algorithmicx}
\usepackage{algpseudocode}

\usepackage{arydshln}
\usepackage{graphicx}
\usepackage{booktabs}
\usepackage{subcaption}  
\usepackage{wrapfig}
\usepackage{booktabs}
\usepackage{array}
\usepackage{multirow}
\usepackage{adjustbox}
\usepackage{lscape}
\usepackage{caption}
\usepackage{wrapfig}
\usepackage[table]{xcolor}
\definecolor{codeblue}{rgb}{0.25, 0.5, 0.5}
\definecolor{codekw}{rgb}{0.35, 0.35, 0.75}
\definecolor{Gray}{gray}{0.95}

\usepackage{pifont}
\usepackage{bbding}

\usepackage{listings}
\lstdefinestyle{Pytorch}{
    language         = Python,
    backgroundcolor  = \color{white},
    basicstyle       = \fontsize{8.0pt}{9pt}\selectfont\ttfamily\bfseries,
    columns          = fullflexible,
    breaklines       = true,
    captionpos       = b,
    commentstyle     = \fontsize{4pt}{4pt}\color{codeblue},
    keywordstyle     = \fontsize{4pt}{4pt}\color{codekw},
    morekeywords     = {with,scatter_,norm,sort},
}

\theoremstyle{plain}

\theoremstyle{definition}

\theoremstyle{remark}

\newlength\savewidth

\newcommand{\cmark}{\ding{51}}

\title{Bit-by-Bit: Progressive QAT Strategy with Outlier Channel Splitting for Stable Low-Bit LLMs}

\author{
\textbf{Binxing Xu$^{1}$\thanks{Equal contribution.}}
\quad
\textbf{Hao Gu$^{2}$\footnotemark[1]}
\quad
\textbf{Lujun Li$^{2}$}
\quad
\textbf{Hao Wang$^{3}$}
\quad
\textbf{Bei Liu$^{2}$}
\\
\textbf{Jiacheng Liu$^{2}$}
\quad
\textbf{Qiyuan Zhu$^{2}$}
\quad
\textbf{Xintong Yang$^{2}$}
\quad
\textbf{Chao Li$^{1}$\thanks{Corresponding authors.}}
\quad
\textbf{Sirui Han$^{2}$\footnotemark[2]}
\quad
\textbf{Yike Guo$^{2}$\footnotemark[2]}
\\[0.5em]
$^{1}$Zhejiang University
\quad
$^{2}$Hong Kong University of Science and Technology
\\
$^{3}$City University of Hong Kong
\\
\texttt{binxing.xu@zju.edu.cn}
\qquad
\texttt{marcusguhao@gmail.com}
}

\begin{document}

\maketitle

\begin{abstract}
Training LLMs at ultra-low precision remains a formidable challenge. Direct low-bit QAT often suffers from convergence instability and substantial training costs, exacerbated by quantization noise from heavy-tailed outlier channels and error accumulation across layers. To address these issues, we present \textsc{Bit-by-Bit}, a progressive QAT framework with outlier channel splitting. Our approach integrates three key components: (1) block-wise progressive training that reduces precision stage by stage, ensuring stable initialization for low-bit optimization; (2) nested structure of integer quantization grids to enable a "train once, deploy any precision" paradigm, allowing a single model to support multiple bit-widths without retraining; (3) rounding-aware outlier channel splitting, which mitigates quantization error while acting as an identity transform that preserves the quantized outputs.
Furthermore, we follow microscaling groups with E4M3 scales, capturing dynamic activation ranges in alignment with OCP/NVIDIA standards. 
To address the lack of efficient 2-bit kernels, we developed custom operators for both W2A2 and W2A16 configurations, achieving up to 11$\times$ speedup over BF16.
Under W2A2 settings, \textsc{Bit-by-Bit} significantly outperforms baselines like BitDistiller and EfficientQAT on both Llama2/3,
achieving a loss of only 2.25 WikiText2 PPL compared to full-precision models.
\end{abstract}

\begin{figure}
    \centering
    \includegraphics[width=1\linewidth]{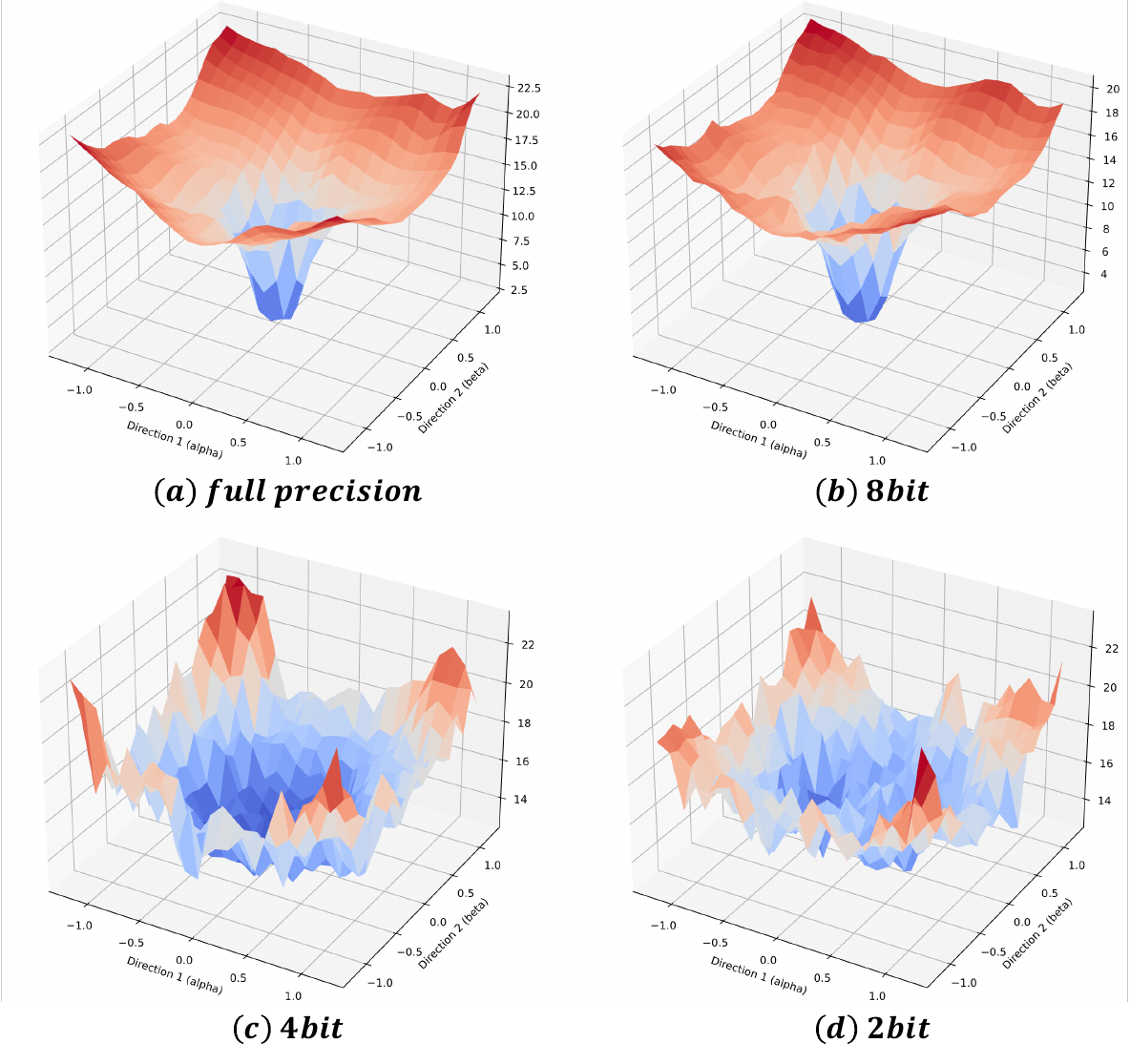}
    \caption{\textbf{Loss landscapes under different precisions.} The vertical axis denotes the loss, the horizontal axes ($\alpha,\beta$) represent random directions in parameter space.}
    \label{fig:loss_landscape}
    \vspace{-4.5mm}
\end{figure}
\section{Introduction}
The remarkable success of modern Large Language Models (LLMs), such as GPT-5 \citep{openai_gpt5} and DeepSeek \citep{liu2024deepseek}, is largely attributed to scaling laws, which suggest that increasing model size consistently enhances performance. However, the burgeoning scale of these models necessitates the adoption of low-precision formats to optimize both storage and computational efficiency. Existing approaches fall into two families: post-training quantization (PTQ) and quantization-aware training (QAT). PTQ quantizes a pretrained model with little or no retraining and thus dominated early work; however, it often degrades sharply at ultralow precisions ($\leq$ 4-bit)~\citep{lin2024awq}. By contrast, QAT incorporates the quantization process directly into the training loop to mitigate the quant error caused by low-precision representation.

\begin{figure*}[t]
    \centering
    \includegraphics[width=1\linewidth]{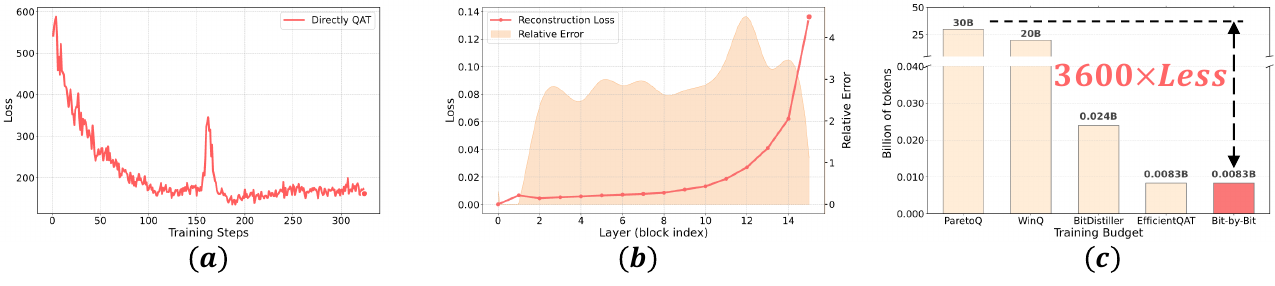}
    \vspace{-5mm}
    \caption{\textbf{Analysis of QAT challenges.} (a) Training loss curve of direct QAT, exhibiting a prominent loss spike. (b) Layer-wise reconstruction loss and relative error across Transformer blocks, illustrating significant error accumulation in deeper layers. (c) Comparison of training budgets; our method (Bit-by-Bit) achieves a 3600$\times$ reduction in token requirements compared to ParetoQ.}
    \label{fig:intro}
    \vspace{-4mm}
\end{figure*}

To ensure low-bit performance, existing QAT methods have explored primarily on several directions: (i) modifying the optimization objective via variants of knowledge distillation~\citep{du2024bitdistiller, chen2024db} to better align with full-precision output distributions; (ii) improving discrete gradient estimation through enhanced Straight-Through Estimators (STE)~\citep{panferov2025quest, malinovskii2024pv} to suppress gradients with large approximation errors; (iii) engineering robust quantization primitives, such as clipping strategies and adaptive grids~\citep{chen2024db, liu2025paretoq, du2024bitdistiller} to mitigate the impact of non-salient values; (iv) employing fine-grained, stage-wise schedules for learning rates and weight decay~\citep{ma2025bitnet, ma2024fbi, team2025minicpm4}; and (v) integrating value-redistributing transformations (e.g., Hadamard) into training~\citep{choi2025rotate, panferov2025quest, tan2025zeroqat, wang2025bitnet} to smooth out outliers prior to quantization.
Despite these advances, existing approaches still face critical stability challenges during low-bit training. They often rely on massive token budgets to converge to usable low-bit representations (Fig.~\ref{fig:intro}(c)); demand extensive hyperparameter “wind tunnel” tuning, particularly of learning rates, since low-bit weights require larger yet inherently unstable updates; and introduce significant computational overhead from complex distillation losses, which slow training and inflate memory usage due to the need to retain both teacher and student logits.
These challenges naturally raise the question: \emph{How can we mitigate quantization error and achieve stable ultra-low-bit QAT?}

\begin{figure*}
    \centering
    \includegraphics[width=1\linewidth]{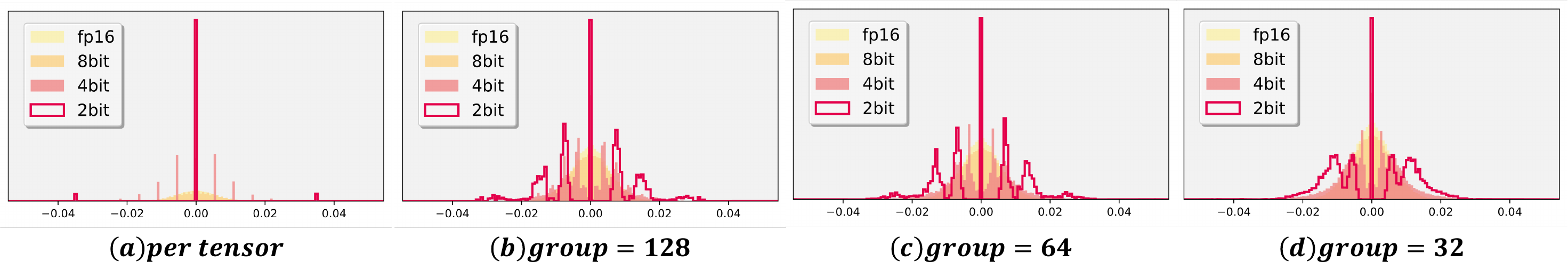}
    \caption{\textbf{Value distributions of various group granularities} showing (a) Low-bit values are nested in the high-bit grid, (b) lower bits collapse representations; larger groups improve dynamic-range.}
    \label{fig:grid}
\end{figure*}

To address these challenges, we first examine the loss landscapes under different precisions (Fig.~\ref{fig:loss_landscape}). We observe that as precision decreases, the loss landscape becomes increasingly uneven and discontinuous, which can trap the model in poor local minima.
Compounding this difficulty, loss spikes emerge during low-bit training process (Fig.~\ref{fig:intro}(a)).
Moreover, 
weight distributions are difficult to represent at low bit widths(Fig.~\ref{fig:grid}), making QAT optimization inherently unstable in the ultra low bit regime. And by further examining the quantization error across different blocks (Fig.~\ref{fig:intro}(b)), we find that later layers suffer from significantly larger errors. This suggests that \emph{the key challenge for ultra-low-bit QAT lies in the accumulation of quantization error}.
So we propose \textbf{Bit-by-Bit}, a progressive framework for stable ultra-low-bit QAT. Our main contributions are:
\begin{itemize}
\item A progressive strategy anneals precision from high to low, quantizing weights first and activations later to provide a well-conditioned start for the subsequent low-bit stage.
\item Extending the curriculum progressive strategy to a unified "any-precision" framework by leveraging the nested nature of bit-width enables "train once, deploy at any precision".
\item Rounding-aware outlier channel splitting, which mitigates both outlier effects and rounding errors while preserving quantized outputs.
\end{itemize}
Our comprehensive evaluation on LLaMA-2/3 and Mistral under both weight-only (w2a16) and weight–activation (w2a2) shows that \textsc{Bit-by-Bit} consistently surpasses strong QAT baselines under the same training budget in ultra–low-bit regimes. On LLaMA-2 7B with w2a2 quantization, it incurs only +2.25 perplexity increase on WikiText2 compared to FP16 (7.72 vs.\ 5.47). Furthermore, on LLaMA-3 family which is hard to quantize, Bit by Bit surpass other QAT methods.

\section{Related work}
\subsection{Quantization for LLMs}
\textbf{Post-Training Quantization (PTQ)} is a mainstream LLM compression method, with aggressive strategies down to 2-bit~\citep{liu2024vptq}, ternary~\citep{kaushal2024spectra}, and binary~\citep{gu2025btc}. Most approaches aim to preserve a small set of salient weights to reduce error, e.g., AWQ~\citep{lin2024awq} uses activation-guided scaling, SqueezeLLM~\citep{kim2023squeezellm} mixes dense/sparse formats, PB-LLM~\citep{shang2023pb} combines binary and INT8, and BiLLM~\citep{huang2024billm} adds residual quantization. Despite effectiveness, these designs often introduce complex implementations and kernel inefficiency.

\textbf{Quantization-Aware Training (QAT)} aims to address these issues by jointly optimizing the weights along with the quantizer to mitigate quantization error, including: LLM-QAT~\citep{liu2023llm} operates without additional data but suffers from high computational overhead during teacher logits computation;
QuEST~\citep{panferov2025quest} filters outlier gradients and employs RMS operations combined with Gaussian and Hadamard transforms for distribution fitting;
DB-LLM~\citep{chen2024db} introduces a dual binary representation along with a deviation-aware distillation loss and BitNet~\citep{ma2025bitnet} has demonstrated the potential of ternary weight representations, yet requires as many as 2T tokens to establish a stable low-bit model.


\textbf{Weight-Only Quantization} stores LLM weights in low precision, with recent works pushing below 1-bit representation~\citep{gu2025btc,dong2024stbllm}, achieving up to $20\times$ compression. \textbf{Weight–Activation Quantization} further quantizes activations, enabling low-precision GEMM kernels and reducing IO (e.g., DeepSeek’s DeepGEMM~\citep{deepseek_deepgemm}). Methods like SmoothQuant~\citep{pmlr-v202-xiao23c} shift quantization difficulty from activations to weights, while rotation-based approaches (QuaRot~\citep{ashkboos2024quarot}, SpinQuant~\citep{liu2024spinquant}) improve robustness via orthogonal transformations. Our QAT framework supports both ultra-low-bit weight-only and weight–activation quantization.

\section{Method}
In this section, we revisit quantization and introduce our method, which integrates a progressive QAT strategy with Once-for-any-precision training, outlier channel splitting, and microscaling groups.
\begin{figure*}
    \centering
    \includegraphics[width=1\linewidth]{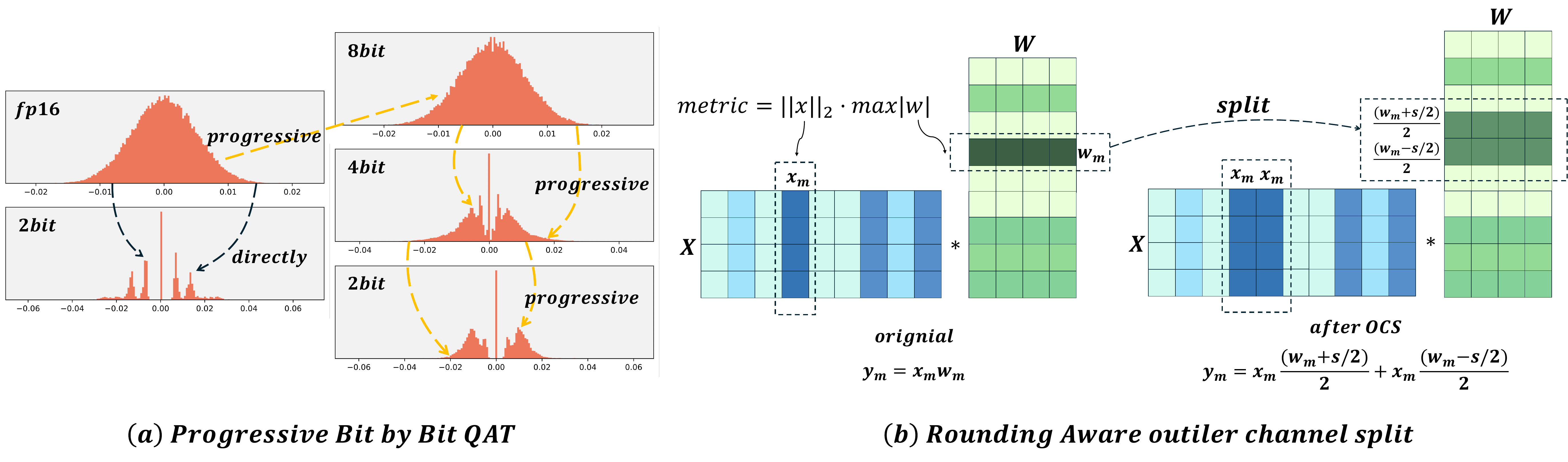}
    \caption{\textbf{(a) Progressive Bit-by-Bit QAT:} Direct 2-bit QAT drives weights into coarse clusters under a non-smooth loss landscape, progressive schedule that lowers precision stage-by-stage, using the higher-precision phase to stabilize and initialize the next stage. \textbf{(b) Rounding-aware outlier channel splitting:} detect outlier channels via metric $||\mathbf{x}||_{2}\cdot max{|w|}$, then apply identical, rounding-aware halving that keeps the quantized output unchanged.}
    \label{fig:main_method}
    \vspace{-4mm}
\end{figure*}
\subsection{Quantization Revisited}
Quantization is applied to all linear layers except the LM head and the embedding layer. In group-wise quantization, weight matrix \(W\in\mathbb{R}^{m\times n}\) is partitioned into \(G=\frac{mn}{g}\) groups of size \(g\):
\begin{align*}
W=\big[\,W^{(1)},W^{(2)},\ldots,W^{(G)}\,\big], W^{(i)}\in\mathbb{R}^{g\times 1}.
\end{align*}
Each group is quantized independently. For any element \(x\in W^{(i)}\), we compute
\[
q=\operatorname{round}\!\left(\frac{x}{s}+z\right),\,
q\leftarrow \operatorname{clip}\!\left(q,\,0,\,2^n-1\right),\,
\]
\[
s=\frac{Max-Min}{2^n-1},\,
z=-\operatorname{round}\!\left(\frac{Min}{s}\right),
\]
where $Max=\max(W^{(i)}), Min=\min(W^{(i)})$. Henceforth, we use the terms scale and step size interchangeably to denote $s$.
Traditional symmetric quantizers are often suboptimal at ultra-low bit-widths.Specifically, under a 2-bit configuration, they either utilize only three distinct levels to maintain a zero-centered balance (e.g., $\{-1, 0, 1\}$), or map weights to a zero-less symmetric codebook such as $\{-1.5, -0.5, 0.5, 1.5\}$ as explored in strategies like SEQ~\citep{liu2025paretoq}. To maximize representation capacity, we adopt an asymmetric quantizer with a zero-point.
To incorporate the quantizer into training, we adopt the straight-through estimator (STE) to address the non-differentiability of the rounding operation during backpropagation.
Gradients flow only through the weights, while the scale $s$ and zero-point $z$ obtained directly from closed-form expressions.
No additional clipping or heuristic adjustment~\citep{shao2023omniquant} is applied to the weights, ensuring a simple yet effective quantization scheme.

\subsection{Progressively Bit-by-Bit QAT}
As shown in Fig.~\ref{fig:loss_landscape} and Fig.~\ref{fig:intro}(a), directly optimizing at very low precision often produces a rugged loss landscape and loss spike, making training to suboptimal local minima.
We observe that dequantized weights at lower bit collapse into limited number of coarse clusters (Fig.~\ref{fig:grid}). Lower-bit values are naturally covered by the higher-precision grid. For any value $x_{\text{low}}$ in the lower-bit grid, there is always a corresponding high-bit value $x_{\text{high}}$ within half a step size $\lvert x_{\text{low}}-x_{\text{high}}\rvert\le \tfrac12 s_{\text{high}}$. This hierarchical relationship suggests a natural coarse-to-fine progress: higher-bit grids act as smooth refinements of lower-bit representations, motivating us to adopt progressive quantization as a more stable optimization scheme.


\textbf{Progressive Strategy.} We begin from a relatively high precision setting, which closely matches full precision and introduces negligible quantization error, providing a well-conditioned initialization. The bitwidth is then gradually reduced across stages (e.g., from 8-bit to 4-bit and finally to 2-bit for weights), allowing the model to progressively adapt to the increasing quantization noise.
For weight–activation quantization, we apply the same principle: the model is first stabilized under a configuration with low-bit weights but high-precision activations, and the activation precision is then progressively lowered in subsequent stages. This staged reduction enables the model to adapt step by step to the growing activation noise, thereby mitigating training instability. We found that reducing weight precision first, followed by activation bits, yields the most stable results, further exploration of alternative strategies is provided in Appendix~\ref{app:activation_progressive}.

\textbf{Block Wise Strategy.} Following BRECQ~\citep{li2021brecq} and EfficientQAT~\cite{chen2024efficientqat}, we employ a block-wise objective to mitigate error accumulation. For block $i$, let $x^{(i)}_{w\mathbf{k}a16}$ denote the input activation when all preceding blocks use $\mathbf{k}$-bit weights (while activations remain FP16), and let $x^{(i)}_{w(\mathbf{k}+\Delta)a16}$ denote the activation obtained when the preceding blocks use a slightly higher precision, e.g., $w4a16$ as $w(2+\Delta)a16$ for stabilizing $w2a16$. The full-precision reference is denoted as $w16a16$. The block-wise loss is formulated as
\begin{align*}
\mathbf{MSE}[ (x^{(i)}_{w(\mathbf{k}+\Delta)a16} W^{(i)}_{w\mathbf{k}a16})-(x^{(i)}_{w16a16} W^{(i)}_{w16a16}) ].
\end{align*}
This design leverages higher-bit block activations as a more accurate teacher, improving the robustness of QAT across 8/4/2-bit regimes. Similar formulation is also applied to weight–activation quantization, where activations are progressively reduced from $a16$ to lower precisions.

\begin{figure}
    \centering
    \includegraphics[width=1\linewidth]{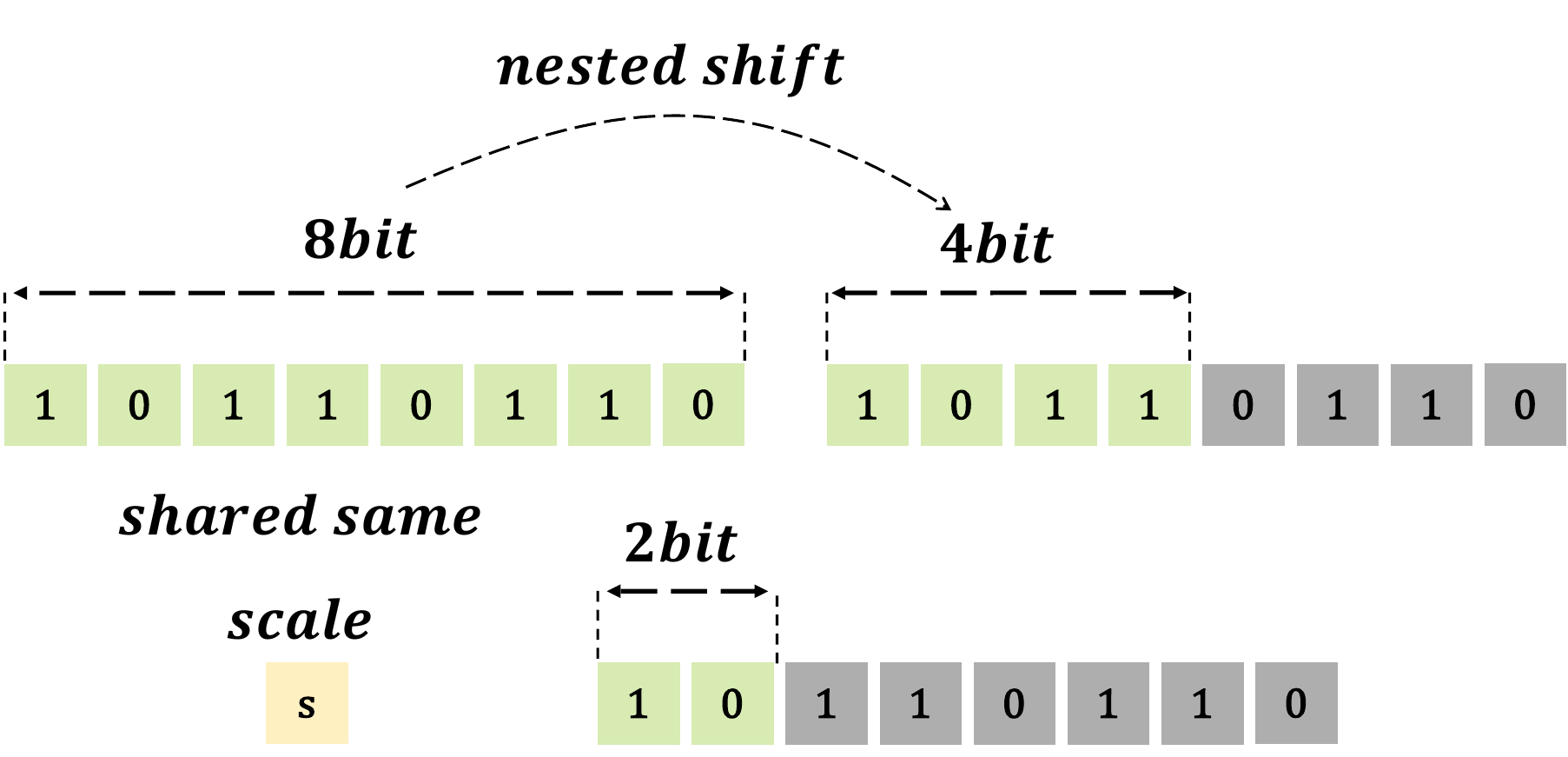}
    \caption{Bit-shifting from higher to lower precision while sharing the same scale $s$.}
    \label{fig:shift}
    \vspace{-3.5mm}
\end{figure}
\subsection{Once-for-any-precision.}
Besides stabilizing low-bit optimization, our progressive strategy also enables a single model to support multiple precisions. Conventionally, supporting multiple bit-widths requires storing several independently trained QAT checkpoints (e.g., W8/W4/W2), incurring considerable training and storage costs. Inspired by~\citep{nair2025matryoshka,park2024anyprecision,cai2019onceforall}, we extend our \textsc{Bit-by-Bit} framework into a unified \textsc{once-for-any-precision} paradigm: a single set of master parameters can be deployed at various bit-widths without additional retraining.

\textbf{Nested low-bit grids via bit shifts.} The key idea is that, \textbf{with the same scale $s$}, a lower-bit quantizer is naturally a coarser version of higher-bit one (e.g., $2\text{bit}\subset 4\text{bit}\subset 8\text{bit}$). If a weight is already quantized to an $h$-bit integer code $q^{(h)}$, we can get an $l$-bit code ($l<h$) by just removing the last $(h-l)$ least significant bits.
\[
q^{(l)}=\Big\lfloor \tfrac{q^{(h)}}{2^{h-l}} \Big\rfloor,
\qquad
\hat{w}^{(l)} = s \cdot \big( q^{(l)} \ll (h-l) \big).
\]
In practice this is just bit shifting as Fig.~\ref{fig:shift}:
\[
q^{(l)} = q^{(h)} \gg (h-l).
\]
\textbf{Curriculum Progressive Strategy.} Leveraging the nested nature of bit-widths, we adopt a curriculum manner from high precision to low precision: higher-bit training provides a well initialize, and lower-bit objectives are added gradually. Concretely, we optimize an expanding set of targets
\[
\texttt{Bit}=\{(8);(8,4);(8,4,2)\},
\]
i.e., we first train only with 8-bit, then jointly with 8/4-bit, and finally with 8/4/2-bit. The resulting objective is
\[
\mathcal{L}=\sum_{b\in \texttt{Bit}} \lambda_{b}\cdot \mathrm{MSE}\!\left(xW_b,\, y\right),
\]
where $W_b$ denotes the shared master weights truncated to $b$ bits, and $\lambda_b$ controls the contribution of each bit-width. \textbf{Deployment.} After training, we keep only the master checkpoint and obtain the desired low-bit on the fly using the nested bit-shift mapping. This enables ``train once, deploy any precision on demand'' without retraining or storing multiple model copies.

\definecolor{codeblue}{rgb}{0.25,0.5,0.5}
\definecolor{codekw}{rgb}{0.85, 0.18, 0.50}

\definecolor{codesign}{RGB}{0, 0, 255}
\definecolor{codefunc}{rgb}{0.85, 0.18, 0.50}

\lstdefinelanguage{PythonFuncColor}{
  language=Python,
  keywordstyle=\color{blue}\bfseries,
  commentstyle=\color{codeblue},  
  stringstyle=\color{orange},
  showstringspaces=false,
  basicstyle=\ttfamily\small,
  literate=
    {*}{{\color{codesign}* }}{1}
    {dataloader}{{\color{codefunc}dataloader}}{1}
    {sample_t_r}{{\color{codefunc}sample\_t\_r}}{1}
    {randn}{{\color{codefunc}randn}}{1}
    {randn_like}{{\color{codefunc}randn\_like}}{1}
    {jvp}{{\color{codefunc}jvp}}{1}
    {stopgrad}{{\color{codefunc}stopgrad}}{1}
    {metric}{{\color{codefunc}metric}}{1}
}

\lstset{
  language=PythonFuncColor,
  backgroundcolor=\color{white},
  basicstyle=\fontsize{9pt}{9.9pt}\ttfamily\selectfont,
  columns=fullflexible,
  breaklines=true,
  captionpos=b,
}
\renewcommand{\thealgorithm}{} 


\begin{algorithm}[t]
\caption{Once-for-any-precision}
\label{alg:once_for_any_precision}
\begin{lstlisting}[language=python]
# Bit: Target bit-width list[[8],[8,4],[8,4,2]]
def once_for_any_precision(M, i, Bit):
    Block = M.blocks[i]
    x_fp, y_fp = Block.data_loader()
    loss_total = 0
    for b in Bit:
        lambda_b = schedule_bit_ratio(b)
        # use bit_shift_mapping
        W_r = s * quantize(Block.W, bits=b)
        y_pred = Block.forward(x_fp, W_r)
        loss_total+=lambda_b * MSE(y_pred, y_fp)
    return loss_total # update

def bit_shift_mapping(s, q_high, h, l):
    q_low = q_high >> (h - l)
    W_low = s * (q_low << (h - l))
    return W_low
\end{lstlisting}
\end{algorithm}

\subsection{Outlier Channel Split}
The outlier issue has long been a major challenge in quantization, for uniform $b$-bit quantization, the step size is $s=\frac{\max(W)-\min(W)}{2^b-1}.$ Weight outliers enlarge the range $R=\max(W)-\min(W)$, thereby increasing $s$; activation outliers enlarge $\|x\|_1$. As a result, the quantization error is bounded by $\big|xW - xW_{\text{quant}}\big|\;\le\;\tfrac{1}{2}s\,\|x\|_1,$ showing that both weight and activation outliers amplify the error through range expansion and input magnitude.
Prior works~\citep{shao2023omniquant} often mitigate this problem by clipping outliers with learnable parameters.
However, outliers value encode important distributional or semantic features~\citep{massive_activation}, and discarding them directly can lead to substantial performance degradation.

Motivated by this, we adopt \textbf{Outlier Channel Splitting (OCS)}~\citep{OCS}. Instead of clipping, OCS duplicates channels containing extreme values and redistributes their contribution via an identity mapping, thereby reducing the dynamic range while preserving critical information.

Consider a linear layer $\mathbf{y} = \mathbf{x} W$. Let $x_m \in \mathbb{R}^{d \times 1}$ denote an identified outlier activation in the $m$-th input channel, and $\mathbf{w}_m \in \mathbb{R}^{1 \times d}$ be its corresponding weight row. OCS splits the original contribution $x_m \mathbf{w}_m$ into two lowered-magnitude branches without changing the numerical output:
\[
x_m \mathbf{w}_m = 
\begin{pmatrix} x_m & x_m \end{pmatrix}
\begin{pmatrix} \tfrac{\mathbf{w}_m+s/2}{2} \\ \tfrac{\mathbf{w}_m-s/2}{2} \end{pmatrix}.
\]
By replacing a single outlier channel with two identical copies of halved magnitude, OCS effectively compresses the dynamic range per channel, which alleviates quantization error at the cost of a minor increase in the input dimension $m$. Further theoretical analysis of the error reduction is provided in Appendix \ref{app:ocs_error}.

Splitting increases layer width and computation, so we split only a small subset of channels. To identify outlier channels that are most susceptible to quantization errors, we introduce a sensitivity metric
$S_i$ for each input channel i. This metric is computed based on statistics gathered from a calibration set. Specifically, For a linear layer with input $\mathbf{x}\in\mathbb{R}^m$ and weights $W\in\mathbb{R}^{m\times n}$, we define an outlier metric for each input channel $i$ as
\[
\mathbf{metric_i} = \|\mathbf{X}_{i}\|_{2} \cdot \max_{1\le j\le n} |W_{ij}|,
\]
where $\|\mathbf{X}_{i}\|_{2}$ denotes the $\ell_2$ norm of the $i$-th input feature aggregated across $N\times L$ tokens, and $\max_{1\le j\le n} |W_{ij}|$ signifies the maximum weight magnitude in the $i$-th channel across all $n$ output dimensions. As shown in Fig.~\ref{fig:intro}(b), quantization error accumulates along depth, later blocks suffer larger errors. Motivated by this observation, we adopt a \emph{block-wise} schedule that linearly increases the split ratio with depth. Index Transformer blocks by $b=1,\dots,B$ from shallow to deep. For block $b$, we set
\[
r_b \;=\; r_{\min} \;+\; \frac{b-1}{B-1}\,\bigl(r_{\max}-r_{\min}\bigr),
\]
and split the top $\lceil r_b\, m\rceil$ input channels (ranked by $s_i$), where $m$ is the number of input channels in that layer. This allocates fewer splits to early blocks and more to later blocks, matching the observed depth-wise error accumulation.

\subsection{Microscaling Format}
Ultra low bit quantization significantly reduces computational and I/O costs, but it also severely restricts the representable dynamic range (Fig.~\ref{fig:grid}). To address this limitation, microscaling formats, such as MXFP4~\citep{rouhani2023microscaling} and NVFP4~\citep{nvidia_nvfp4_2025}, introduce a shared scale factor applied to small blocks of weights. Follow these line, we apply per-group scaling over 32 elements and store each group scale in FP8 to minimize overhead. While standard MX formats adopt E8M0 (power-of-two) scaling, this approach is not granular enough for 2-bit models. We use E4M3 FP8 for group scales instead. This format provides sufficient mantissa precision for accurate step-size adjustment, while adding only one 8-bit scale per 32 weights, resulting in a storage overhead of just $8/32 = 0.25$ bits per weight.

\section{Experiment}
We comprehensively evaluate \textbf{Bit-by-Bit} against both post-training quantization (PTQ) and quantization-aware training (QAT) baselines. PTQ methods include GPTQ~\citep{frantar2022gptq}, AWQ~\citep{lin2024awq}, OmniQuant~\citep{shao2023omniquant}, SmoothQuant~\citep{xiao2023smoothquant}, MatQuant~\citep{nair2025matryoshka}, and SpinQuant~\citep{liu2024spinquant}, while QAT baselines cover EfficientQAT~\citep{chen2024efficientqat}, ParetoQ~\citep{liu2025paretoq}, and BitDistiller~\citep{du2024bitdistiller}. All experiments are run on a single H800 GPU.

\begin{table*}[t]
\centering
\caption{Evaluation results on WikiText2 and C4 across different model sizes. Our method \textbf{Bit-by-Bit} is highlighted.}
\label{tab:main_result}
\setlength{\tabcolsep}{4pt}
\renewcommand{\arraystretch}{1.25}
\resizebox{\textwidth}{!}{%
\begin{tabular}{lcc rrrr rrrr}
\toprule
\multirow{2}{*}{\textbf{Method}} & \multirow{2}{*}{\textbf{Bits}} & \multirow{2}{*}{\textbf{Group}} 
& \multicolumn{4}{c}{\textbf{WikiText2}} & \multicolumn{4}{c}{\textbf{C4}} \\
\cmidrule(lr){4-7} \cmidrule(lr){8-11}
 & & & 2-7B & 3.2-1B & 3.2-3B & 3-8B & 2-7B & 3.2-1B & 3.2-3B & 3-8B \\
\midrule
\rowcolor{gray!8}
FP16 & - & - & 5.47& 9.75& 7.81& 6.13& 6.97& 12.74& 10.44& 8.89\\
\midrule
\rowcolor{cyan!15}
\multicolumn{11}{c}{\textbf{Weight Only Quantization (w2a16)}} \\
\midrule
GPTQ         & w2a16 & 32 & 60.5& 2775.63& 379.23& 43.34& 33.7& 1875.41& 323.24& 43.28\\
AWQ          & w2a16 & 32 & 2.2e5& 1.7e7& 7.2e6& 5.2e5& 1.75e5& 1.9e7& 7.7e6& 5.1e5\\
OmniQuant    & w2a16 & 32 & 11.06& 6260.71& 1.4e51& 2.2e6 & 15.02& 2442.55& 8315.17& 8.3e5\\
ParetoQ      & w2a16 &  -1  & 10.89& 42.82& 26.88& 100.04& 12.40& 35.08& 24.08& 94.97\\
EfficientQAT & w2a16 & 32 & 7.39& 21.48& 13.31& 11.17& 9.30& 24.84& 17.38& 15.18\\
BitDistiller & w2a16 & 32 & 7.28& 20.41& 12.80& 10.40& 10.01& 31.24& 19.86& 18.23\\
\rowcolor{yellow!15}
\textbf{Bit-by-Bit (Ours)} & w2a16 & 32 & \textbf{6.50}& \textbf{16.13}& \textbf{11.02}& \textbf{8.32}& \textbf{9.22}& \textbf{23.03}& \textbf{16.45}& \textbf{14.27}\\
\midrule
\rowcolor{cyan!15}
\multicolumn{11}{c}{\textbf{Weight Activation Quantization (w2a2)}} \\
\midrule
SmoothQuant  & w2a2& 32 & 2.5e5& 1.7e7& 2.0e6& 8.6e6& 3.0e5& 1.8e8& 1.5e6& 9.9e6\\
SpinQuant    & w2a2& 32 & 5433.06& 4059.73& 4008.33& 7931.37& 7524.73& 8222.23& 8256.53& 1.3e5\\
ParetoQ      & w2a2&  -1  & 259.74& 1091.78& 1018.61& 549.71& 135.32& 418.22& 401.22& 237.21\\
EfficientQAT & w2a2& 32 & 9.71& 29.42& 20.19& 17.93& \textbf{10.89}& 66.53	& 31.65& 26.58\\
BitDistiller & w2a2& 32 & 29.66& 30.68& 18.39& 15.36& 43.08& 60.12& 28.23& 25.86\\
\rowcolor{yellow!15}
\textbf{Bit-by-Bit (Ours)} & w2a2& 32 & \textbf{7.72}& \textbf{22.71} & \textbf{13.87}& \textbf{11.51}& 12.87& \textbf{46.53} & \textbf{23.63}& \textbf{21.58}\\
\bottomrule
\end{tabular}
}
\end{table*}
\subsection{Experimental Settings}
We test on the LLaMA~\citep{dubey2024llama} and Mistral families, evaluating five zero-shot reasoning benchmarks (PIQA, ARC-Easy, ARC-Challenge, HellaSwag, Winogrande) and two language modeling tasks (WikiText2~\citep{merity2016pointer} and C4~\citep{raffel2020exploring}).


For PTQ baselines, we use a 256-sample RedPajama subset (seq length 2048) for AWQ, GPTQ, and SmoothQuant; OmniQuant follows its 40-epoch calibration, and SpinQuant is calibrated for 2 epochs. For QAT baselines, EfficientQAT adopts Block-AP (4096 RedPajama samples, 2 epochs) followed by E2E on Alpaca; BitDistiller uses a 4096-sample Alpaca subset for KD-based QAT; and ParetoQ is trained on 4096 RedPajama + 4096 Alpaca samples for 2 epochs, aligned to our budget (vs. 30B tokens in the original). Since these methods target weight-only quantization, we extend them with activation quantizers: online dynamic scaling for EfficientQAT, asymmetric clipping for BitDistiller, and 2-bit SEQ for ParetoQ.
We train Bit-by-Bit on a 4096-sample subset of RedPajama. For weight-only quantization, the model precision is progressively reduced from \texttt{w8a16} to \texttt{w4a16} and then to \texttt{w2a16}, switching every two epochs, while splitting 10\% of weight channels as detected by the metric. For weight–activation quantization, we first lower the weight precision to \texttt{w2a16}, then reduce the activation precision to \texttt{w2a2} progressively.

\begin{table}[t]
    \centering
    \caption{Zero-shot evaluation of LLaMA-3.2 3B on five downstream tasks accuracy.}
    \resizebox{0.48\textwidth}{!}{
    \begin{tabular}{cc|ccccc|c} 
    \toprule[\heavyrulewidth]
        \multicolumn{2}{c|}{\textbf{LLaMA-3.2-3B}} & PIQA & Hella. & Wino. & ARC-c & ARC-e & Avg \\ 
        \midrule \midrule
        \rowcolor{gray!8}
        \multicolumn{2}{c|}{bfloat16} & 77.47 & 73.62 & 69.61 & 45.90 & 71.71 & 67.67 \\ 
        \midrule
        ~ & ParetoQ & 66.70 & 43.48 & 52.49 & 21.93 & 44.36 & 45.79\\ 
        \multirow{2}{*}{w2a16} & EfficientQAT & 70.02 & 57.07 & 59.35 & 34.13 & 58.92 & 55.89\\ 
        
        & BitDistiller & 70.65 & 57.42 & 59.78 & 34.71 & 58.34 & 56.18\\ 
        ~ & \multicolumn{1}{>{\columncolor{yellow!15}}c|}{\textbf{Bit-by-Bit (ours)}} 
          & \multicolumn{1}{>{\columncolor{yellow!15}}c}{71.87} 
          & \multicolumn{1}{>{\columncolor{yellow!15}}c}{58.03} 
          & \multicolumn{1}{>{\columncolor{yellow!15}}c}{60.38} 
          & \multicolumn{1}{>{\columncolor{yellow!15}}c}{35.58} 
          & \multicolumn{1}{>{\columncolor{yellow!15}}c}{58.71} 
          & \multicolumn{1}{>{\columncolor{yellow!15}}c}{\textbf{56.91}} \\ 
        \midrule
        ~ & ParetoQ & 51.80 & 25.76 & 48.78 & 23.55 & 27.53 & 35.48 \\ 
        \multirow{2}{*}{w2a2} & EfficientQAT & 56.53 & 34.76 & 52.17 & 21.84 & 35.23 & 40.10\\ 
        & BitDistiller & 60.87 & 42.15 & 54.03 & 26.72 & 47.61 & 46.28\\ 
        ~ & \multicolumn{1}{>{\columncolor{yellow!15}}c|}{\textbf{Bit-by-Bit (ours)}} 
          & \multicolumn{1}{>{\columncolor{yellow!15}}c}{66.00} 
          & \multicolumn{1}{>{\columncolor{yellow!15}}c}{49.30} 
          & \multicolumn{1}{>{\columncolor{yellow!15}}c}{56.91} 
          & \multicolumn{1}{>{\columncolor{yellow!15}}c}{31.40} 
          & \multicolumn{1}{>{\columncolor{yellow!15}}c}{54.00} 
          & \multicolumn{1}{>{\columncolor{yellow!15}}c}{\textbf{51.52}} \\ 
    \bottomrule
    \end{tabular}}
    \label{tab:zero_shot}
\end{table}

\subsection{Main Results}
Table~\ref{tab:main_result} reports perplexity results on WikiText2 and C4 under both weight-only (w2a16) and weight-activation (w2a2) settings. \textbf{Bit-by-Bit} consistently surpasses ParetoQ, EfficientQAT, and BitDistiller across model sizes and datasets. In w2a16, it requires fewer training tokens than ParetoQ, converges faster than BitDistiller, and achieves more stable training than EfficientQAT, e.g., reaching 11.02/16.45 PPL on WikiText2/C4 with LLaMA-3.2 3B. The advantage is also pronounced in w2a2, where it reduces WikiText2 PPL on LLaMA-2 7B to 7.72. Zero-shot results (Table~\ref{tab:zero_shot}) further confirm its robustness: Bit-by-Bit achieves the best average accuracy under both w2a16 (56.91) and w2a2 (51.52), exceeding the strongest baseline by over 5 points in the latter. These results demonstrate Bit-by-Bit’s effectiveness in preserving strong generalization under ultra-low precision.

\subsection{Once-for-any-precision evaluation}
Our \textsc{once-for-any-precision} method produces models at multiple bit-widths. To validate the generality of this approach, we compare against MatQuant~\citep{nair2025matryoshka} and OmniQuant~\citep{shao2023omniquant} on Mistral-7B. Specifically, we perform a single QAT run with Bit-by-Bit and directly apply the trained model to different bit-widths (w8a16, w4a16, w2a16). In contrast, the baseline OmniQuant requires separate training for each bit-width, while MatQuant also employs a one-shot QAT strategy for multi-bit adaptation. As shown in Table~\ref{tab:mistral_multi_bit}, our method achieves competitive results under all settings. For w8a16 and w4a16, Bit-by-Bit matches the full-precision baseline with only marginal degradation. obtaining task averages of 73.51 and 73.21 respectively.
In challenging w2a16 setting, Bit-by-Bit excels in w2a16 (65.37 avg/10.73 PPL), surpassing OmniQuant and paralleling MatQuant. This demonstrates that a single QAT process suffices for flexible deployment, eliminating the need for separate retraining.

\newcommand{\ocsmetric}{$||\mathbf{x}||_{2}\cdot \max{|w|}$}

\begin{table*}[!ht]
    \caption{Ablation study on Llama 3.2-1b on w2a16 setting.\label{tab:ablation}}
    \centering
    \resizebox{1.0\textwidth}{!}{
    \begin{tabular}{cccccccc}
        \toprule
        \rowcolor{gray!15}
        Block-wise & Progressive & Ocs & Metric & group size & WikiText2 ppl & Task avg & Memory \\ 
        \midrule
        -      & -         &   -    &    -            & 32  & 1.7e3 & 35.09 & 0.33GB \\
        \cmark & -         &   -    &    -            & 32  & 31.88 & 40.87  & 0.33GB \\
        \cmark & \cmark      &   -    &    -            & 32  & 24.60 & 43.26  & 0.33GB \\
        \cmark & \cmark      & \cmark & Kurtosis        & 32  & 22.43 & 43.69  & 0.36GB \\
        \cmark & \cmark      & \cmark & $w_{\max}$      & 32  & 20.37 & 44.26  & 0.36GB \\
        \cmark & \cmark      & \cmark & $x_{\max}$      & 32  & 19.07 & 44.30  & 0.36GB \\
        \cmark & \cmark      & \cmark & \ocsmetric      & 32  & 17.07 & 45.18  & 0.36GB \\
        \addlinespace[0.2em]
        \hdashline
        \addlinespace[0.2em]
        \cmark & \cmark      & \cmark & \ocsmetric      & 64  & 30.26 & 40.66  & 0.34GB \\
        \cmark & \cmark      & \cmark & \ocsmetric      & 128 & 38.92 & 38.60  & 0.32GB \\
        \bottomrule
    \end{tabular}
    }
\end{table*}

\begin{table}[!ht]
\centering
\caption{Evaluation of Mistral-7B under different quantization methods}
\label{tab:mistral_multi_bit}
\small 
\setlength{\tabcolsep}{5pt} 
\begin{tabular}{clcc} 
\toprule
\multicolumn{4}{c}{\textbf{Mistral-7B}} \\
\textbf{Bits} & \textbf{Method} & \textbf{C4 ppl} & \textbf{Task avg} \\
\midrule
\rowcolor{gray!8}
bfloat16 &   & 8.24 & 73.99\\
\midrule
\multirow{3}{*}{w8a16} 
  & OmniQuant          & 8.24  & 73.77 \\
  & MatQuant          & 8.43  & 73.46 \\
  \rowcolor{yellow!15} & \textbf{Bit-by-Bit (ours)} & 8.33 & 73.51 \\
\midrule
\multirow{3}{*}{w4a16} 
  & OmniQuant          & 8.47  & 73.62 \\
  & MatQuant          & 8.63  & 73.13 \\
  \rowcolor{yellow!15} & \textbf{Bit-by-Bit (ours)} & 8.79 & 72.21 \\
\midrule
\multirow{3}{*}{w2a16} 
  & OmniQuant          & 50.99 & 59.74 \\
  & MatQuant          & 13.05 & 65.99 \\
  \rowcolor{yellow!15} & \textbf{Bit-by-Bit (ours)} & 10.73 & 65.37 \\
\bottomrule
\end{tabular}
\end{table}
\subsection{Ablation}
We conduct a comprehensive ablation of our proposed components on LLaMA3.2-1B. As shown in Table~\ref{tab:ablation}, using block-wise loss yields substantially better results than end-to-end training with NLL loss. Training directly on w2a16 performs poorly, whereas adopting progressive training improves convergence and accuracy. Incorporating OCS brings further gains. We evaluate several metrics for detecting outlier channels, including weight maximum ($w_{\max}$), activation maximum ($x_{\max}$), and kurtosis~\citep{decarlo1997meaning,nrusimha2024mitigating} which measures the “tailedness” of distribution, and find that the combined weight–activation metric $\|\mathbf{x}\|_{2} \cdot \max|{w}|$ yields the best performance. While OCS slightly widens the weight matrix, the memory overhead remains modest (0.33GB$\rightarrow$0.36GB).
About the impact of groupsize: using group-128 saves only 0.04GB of memory but leads to a sharp degradation in performance where task accuracy falls from 45.18 to 38.60. Furthermore, we provide additional ablation results for the w2a2 configuration in Appendix~\ref{app:w2a2_ablation}.

\begin{figure}
    \centering
    \includegraphics[width=1\linewidth]{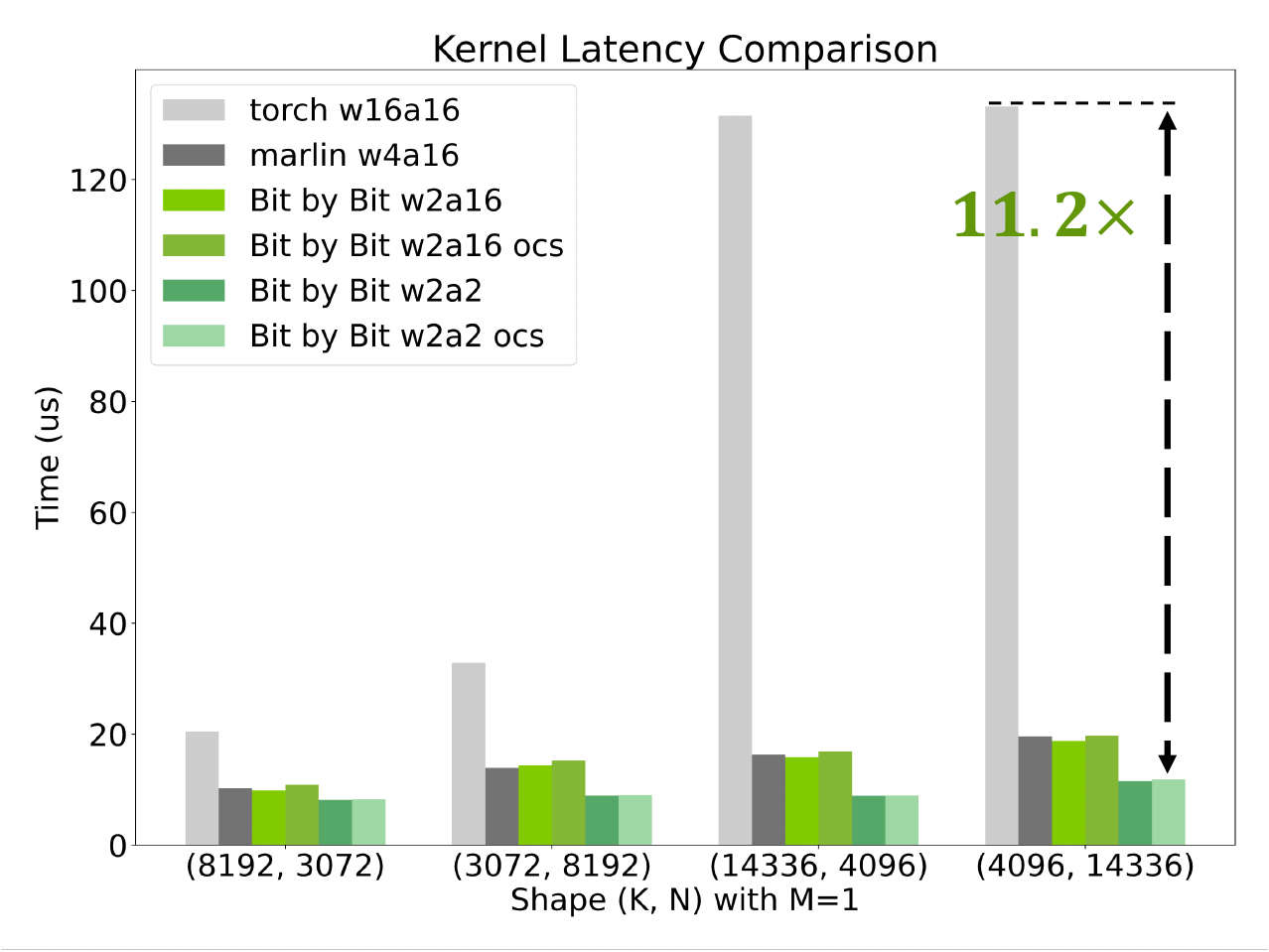}
    \caption{Kernel latency of \textsc{Bit-by-Bit} relative to native PyTorch and Marlin of $W_{\text{up}}$, $W_{\text{down}}$.}
    \label{fig:speed}
    \vspace{-3mm}
\end{figure}
\subsection{Speed Measurement}
Modern GPU architectures are highly optimized for standard precision types (e.g., FP16, BF16) and specific integer formats (INT8/INT4).
To address the lack of native 2-bit support on modern GPUs, we implement specialized high-performance CUDA kernels for W2A16 and W2A2 GEMV operations.

As shown in Fig.~\ref{fig:speed}, the y axis represents the latency of a GEMV operation between an $(1, K)$ input vector $x$ and a $(K, N)$ weight matrix $W$. The x axis denotes matrix shapes corresponding to $W_{\text{down}}$ and $W_{\text{up}}$ from MLP layers of Llama 3.2-3B and Llama 3-8B. Results reported with ocs are measured on weight matrices that have been expanded to accommodate the additional channels introduced by the outlier splitting process.
Notably, in $(4096, 14336)$ setting, our W2A2 implementation achieves a speedup of over $10\times$ compared to the native PyTorch FP16 baseline, the performance overhead remains negligible with the inclusion of OCS. Furthermore, for end-to-end inference on Llama 3-8B, we reaches a decoding throughput of 76 tokens/s, representing a 1.5$\times$ speedup over the 49 tokens/s of Transformers baseline. Detailed implementation of our custom kernels, performance results and test setting are provided in Appendix~\ref{app:operator}.


\section{Conclusion}
We introduced \textsc{Bit-by-Bit}, a stable low bit QAT framework integrating block-wise progressive precision schedule, a once-for-any-precision multi target objective, and rounding aware outlier-channel splitting that preserves the quantized output while shrinking rounding error. By treating low-bit training as a coarse to fine adaptation, it achieves stable convergence and enables flexible multi-precision deployment from a single trained model.
\bibliography{acl2026}
\newpage
\appendix
\section*{Appendix}

\section{Extended Discussion}
\subsection{The Use of Large Language Models (LLMs)}
A large language model was utilized for grammatical and stylistic refinement of the manuscript. Its role was strictly limited to text editing and polishing to enhance clarity. All research ideas, experimental design, and analytical content are the original work of the authors.

\subsection{Broader Impacts}
Our work advances ultra-low-bit quantization of large language models through a progressive training strategy with outlier channel splitting. By enabling stable training at 2-bit and below, \textbf{Bit-by-Bit} reduces the memory footprint and computational cost of LLMs by orders of magnitude. This improvement directly translates into lower inference latency, reduced energy consumption, and smaller carbon emissions, making the deployment of LLMs more sustainable. 

Beyond efficiency, democratization is another key impact: with drastically reduced hardware requirements, powerful LLMs become accessible to a wider range of users and organizations, including those with limited computing resources. This may empower broader participation in AI research and applications, bridging the gap between well-funded institutions and smaller labs or industry players. 

On the societal side, compressed LLMs can be deployed in edge scenarios such as mobile devices, offline environments, and privacy-sensitive settings, expanding the reach of AI to education, healthcare, and accessibility applications. However, lowering the barriers to deployment also amplifies risks of misuse, such as generating disinformation at scale or enabling harmful applications on inexpensive hardware. Mitigating these risks requires complementary safeguards, responsible governance, and continued community awareness. 

Overall, we believe our work contributes to the ongoing effort of making LLMs greener, more efficient, and more inclusive, while highlighting the importance of balancing technological progress with responsible use.

\subsection{Limitations}
While \textsc{Bit-by-Bit} improves stability at ultra–low bits, it has several limitations. (i) We observe larger performance drops on the Qwen family, these models appear harder to quantize, leading to greater quantization error, and a deeper analysis is left for future work. (ii) The block-wise training schedule is less friendly to distributed training than end-to-end schemes, requiring nontrivial load-balancing and communication engineering. (iii) We have not extensively explored direct end-to-end progressive training; its convergence behavior and trade-offs remain open. (iv) We have not explored directions include learning layerwise schedules and split ratios automatically, extending to MoE and longer-context inference (e.g., KV-cache quantization), integrating hardware-aware mixed-precision search, and combining our training with lightweight distillation.

\subsection{Ethics statement}
We acknowledge and adhere to the ACL Code of Ethics. We have carefully considered the ethical implications of our research and paper submission. Our work does not involve human subjects, and it does not make use of data sets that could raise privacy or security concerns. We have ensured that our methodology and applications do not introduce or perpetuate harmful biases, and we have taken care to document our data sources and experimental procedures to promote transparency and reproducibility. We have no known conflicts of interest or sponsorship to disclose.

\subsection{Reproducibility statement}
We are committed to providing sufficient detail for the academic community to reproduce the results presented in this paper.
All experiments were performed on NVIDIA H800 GPU. We utilized the official implementations of all baseline methods where available, ensuring consistent environment configurations. Our evaluations were conducted on two major model families: the LLaMA series and the Mistral series. Performance was measured across seven standard benchmarks: Zero-Shot Reasoning: PIQA, ARC-Easy, ARC-Challenge, HellaSwag, and Winogrande; Language Modeling: WikiText2 and the C4 test set.
We took measures to align the training cost across all QAT approaches for an unbiased evaluation.
- EfficientQAT was first subjected to the Block-AP stage, utilizing a 4096-sample RedPajama subset over 2 epochs, and then proceeded to the E2E stage using the entire Alpaca dataset.
- For BitDistiller, knowledge distillation was performed on a 4096-sample Alpaca subset synthesized by the teacher model.
- ParetoQ’s training budget was limited to 2 epochs, leveraging a combined dataset comprising a 4096-sample RedPajama subset and an equal-sized 4096-sample Alpaca subset.
Furthermore, because these QAT baselines were inherently weight-only, we customized the activation quantization for each: EfficientQAT used a dynamic quantizer, BitDistiller relied on asymmetric clipping, and ParetoQ was equipped with a 2-bit SEQ quantizer.
We used a 4096-sample subset of RedPajama in our Bit-by-Bit training process. In the process of Weight-Only Quantization, we incorporated the splitting of 10\% of weight channels based on the metric at each step. In the process of Weight-Activation Quantization, we maintain the 10\% channel splitting rule.

\section{Extended and detail Method}

\subsection{Different Progressive Strategies}\label{app:activation_progressive}
\subsubsection{Precision Progressive Strategies}
\paragraph{(A) Weights $\rightarrow$ Activations (claimed in method).}
We first lower the \emph{weight} precision to stabilize the network under weight noise, and only then reduce the \emph{activation} precision:
\begin{align*}
&(w8,a16)\;\rightarrow\;(w4,a16)\;\rightarrow\;(w2,a16)\\
&\rightarrow\;(w2,a8)\;\rightarrow\;(w2,a4)\;\rightarrow\;(w2,a2).
\end{align*}
\paragraph{(B) Activations $\rightarrow$  Weights.}
First lower the \emph{activation} precision then reduce the \emph{weights} precision:
\begin{align*}
&(w16,a8)\;\rightarrow\;(w16,a4)\;\rightarrow\;(w16,a2)\\&\rightarrow\;(w8,a2)\;\rightarrow\;(w4,a2)\;\rightarrow\;(w2,a2).
\end{align*}
\paragraph{(C) Alternating W/A.}
We interleave the bit reductions of weights and activations:
\begin{align*}
&(w8,a16)\;\rightarrow\;(w8,a8)\;\rightarrow\;(w4,a8)\\&\rightarrow\;(w4,a4)\;\rightarrow\;(w2,a4)\;\rightarrow\;(w2,a2).
\end{align*}
\paragraph{(D) Cyclic Precision~\citep{kim2022ctmq}}
Unlike monotone schedules, cyclic precision alternates between $(k{+}1)$- and $k$-bit training before committing to $k$-bit. The idea is to leverage the smoother loss landscape of $(k{+}1)$-bit to recalibrate scales and reduce STE bias, while gradually adapting to the coarser $k$-bit lattice. A typical sequence is
\begin{align*}
(w16,a16)\!\to\!(w3,a16)\!\to\!(w2,a16)\!\to\\(w3,a16)\!\to\!(w2,a16)\cdots\to(w2,a2).
\end{align*}

In practice, we first warm up from 8-bit down to $(k{+}2)$-bit, then run several short cycles between $(k{+}1)$ and $k$, and finally fine-tune at $k$-bit. This cyclic back-and-forth helps avoid representation collapse at ultra-low bits (e.g., 2-bit) by ensuring parameters remain quantizable on both lattices. While it introduces extra bit switches and hyperparameters, it often improves stability compared to a one-shot drop.

\paragraph{Empirical observations.}
We typically find Schedule (A) more stable (smoother loss/PPL decay, fewer divergence events), likely because it avoids simultaneous large shifts in both parameter and activation distributions. The alternating scheme can work but is more sensitive to optimizer and clipping hyperparameters and often requires longer warmup.

\subsubsection{Block-wise Progressive Strategy}
We adopt a stochastic, depth-aware curriculum over transformer blocks. Let the model have $L$ blocks indexed from input to output as $j=1,\dots,L$. At stage $t$ (with target bit $b_t$), we quantize only a subset $\mathcal{S}_t\subseteq\{1,\dots,L\}$, sampled with a bias toward earlier blocks and with an increasing coverage over stages.

\paragraph{Depth-biased sampling.}
Define a per-block sampling probability
\[
p_j \;\propto\; (L+1-j)^\alpha,\qquad \alpha\!\ge\!0,
\]
so earlier blocks (small $j$) are more likely to be selected. Given a stage-wise coverage ratio $r_t\!\in\!(0,1]$, we sample $|\mathcal{S}_t|=\lfloor r_t L\rfloor$ blocks without replacement according to $\{p_j\}$.

\paragraph{Bit schedule.}
We follow a high-to-low bit curriculum, e.g.,
\[
b_1=8 \;\rightarrow\; b_2=4 \;\rightarrow\; b_3=2,
\]
and optionally apply the same scheme to activations after weights. The coverage ratio increases with $t$ (e.g., $r_t$ linear or cosine from $r_1\!\approx\!0.3$ to $r_T\!=\!1.0$).

\begin{algorithm}[t]
\caption{Block-wise Progressive Strategy\label{app:blockwise}}
\begin{algorithmic}[1]
\State \textbf{Input:} blocks $1..L$, stages $t=1..T$, bits $\{b_t\}$, ratios $\{r_t\}$, bias $\alpha$
\For{$t=1$ \textbf{to} $T$} \Comment{progressively lower precision}
    \State Compute $p_j \propto (L{+}1{-}j)^\alpha$ and sample $\mathcal{S}_t$ with $|\mathcal{S}_t|=\lfloor r_t L\rfloor$
    \For{$j=1$ \textbf{to} $L$}
        \If{$j \in \mathcal{S}_t$} \State Quantize block $j$ to bit $b_t$; \hspace{0.3em}\textit{(others stay at previous bit)}
        \EndIf
    \EndFor
    \State (Optional) apply OCS to top-$r_\ell$ channels in selected blocks
    \State QAT for a fixed budget (steps/epochs) with short LR warmup
\EndFor
\end{algorithmic}
\end{algorithm}

\paragraph{Notes.}
(1) Depth bias ($\alpha$) and coverage growth ($r_t$) control stability/speed; we find $\alpha\!\in[0.5,1]$ and linear $r_t$ robust. 
(2) This stochastic schedule avoids large simultaneous distribution shifts and is more kernel-friendly than fully per-step rebitting. 
(3) For a deterministic variant, select the first $\lfloor r_t L\rfloor$ blocks at each stage instead of sampling.

\subsection{Mixed-precision of down-projection}
As observed by~\citep{chen2025scaling}, the inputs to the MLP down-projection (\textit{FC2 Proj}) in Transformer blocks exhibit persistent activation outliers (high kurtosis). Under ultra–low-bit W/A quantization (e.g., W2A2), these heavy tails dominate the activation quantization error. To remove this bottleneck, we adopt a \emph{layer-wise mixed-precision} scheme that raises the activation bit-width only for outlier-dominated sites while keeping the rest of the network at low precision. Concretely, we compute per-layer activation kurtosis $\kappa$ on a calibration set and mark layers with $\kappa>\tau$ as outlier-sensitive; for these layers we set $w2a4$ (with the same group-wise scaling as elsewhere), while all remaining layers use $w2a2$. This targeted relaxation substantially reduces activation quantization error—especially at coarse group sizes—while incurring minimal overhead and preserves the benefits of ultra–low-bit quantization in the rest of the model.

\subsection{LoRA for Distribution-Preserving Progression}
As illustrated in Fig.~\ref{fig:main_method} (a), the higher-bit stage establishes a well-conditioned weight/activation distribution that serves as a strong initialization for subsequent lower-bit stages. To preserve this distribution while reducing precision progressively, we insert low-rank adapters (LoRA)~\citep{hu2022lora} and restrict updates to these adapters rather than the full quantized backbone.

Concretely, when moving from bitwidth $b_t$ to $b_{t+1}$ ($b_{t+1}<b_t$), we freeze the backbone weights $W^{(t)}$ and optimize only a rank-$r$ perturbation
\begin{align*}
&W^{(t+1)} \;=\; W^{(t)} \;+\; \alpha\, A^{(t)} {B^{(t)}}^\top,\\
&A^{(t)}\in\mathbb{R}^{d\times r},\; B^{(t)}\in\mathbb{R}^{k\times r},  
\end{align*}

with the forward pass quantized as
\begin{align*}
W^{(t+1)}_{\!q}\;=\;Q_{s^{(t+1)}}\!\big(W^{(t)}+\alpha\,A^{(t)}{B^{(t)}}^\top\big).
\end{align*}
To further stabilize the transition, we use a light distribution-matching regularizer that anchors first/second-order statistics of either weights or activations across stages, e.g.,
\begin{align*}
\mathcal{L}_{\text{dist}}
=\big\|\mu(W^{(t+1)}_{\!q})-\mu(W^{(t)}_{\!q})\big\|_2
\\
+\lambda\big\|\sigma(W^{(t+1)}_{\!q})-\sigma(W^{(t)}_{\!q})\big\|_2,
\end{align*}
optionally combined with a KL term on layer activations. In practice we adopt small ranks ($r\in\{4,8\}$) and reinitialize adapters at each stage. This \emph{distribution-preserving} LoRA update significantly mitigates representation drift and reduces instability at ultra-low bits (e.g., 2-bit), while cutting trainable parameters to a $\tfrac{r(d+k)}{dk}$ fraction of full fine-tuning. After convergence, adapters are merged and requantized or discarded after re-estimating scales.

\subsection{Symmetric Microscaling via SEQ}
Our main pipeline uses asymmetric integers for simplicity, whereas microscaling formats (e.g., MXFP4/NVFP4) favor \emph{symmetric} payloads with zero-point fixed at $0$. To avoid the 2-bit degeneration to ternary under strict symmetric uniform grids, we adopt \emph{Stretched Elastic Quantization (SEQ)}~\citep{liu2025paretoq}, an LSQ-style amendment tailored for low-bit settings.\[
W_{\!Q}
=\alpha\!\left(\frac{\big\lfloor \mathrm{Clip}\!\big(\tfrac{W}{\alpha},-1,1\big)\cdot \tfrac{k}{2}-\tfrac{1}{2}\big\rfloor+\tfrac{1}{2}}{k}\right)\times 2,
\]
which places centers at half-integers; for \(b{=}2\) the normalized levels are \(\{-\tfrac{3}{4},-\tfrac{1}{4},\tfrac{1}{4},\tfrac{3}{4}\}\).
Here \(\alpha\in\text{FP8}\) is stored/rounded in FP8 per group, and \(S_{\mathrm{T}}\in\text{FP32}\) is shared per tensor. The dequantized values are
\begin{align*}
&\hat{W}=S_{\mathrm{T}}\cdot W_{\!Q}
= S_{\mathrm{T}}\,\alpha\cdot\Big(n+\tfrac{1}{2}\Big)
\\
&n\in\Big\{-\tfrac{k}{2},\ldots,\tfrac{k}{2}-1\Big\}.
\end{align*}

At \(b{=}2\), the LUT becomes
\[
\mathcal{C}_{\text{SEQ-2b}}=S_{\mathrm{T}}\,\alpha\cdot\{-1.5,-0.5,0.5,1.5\}.
\]
This keeps a zero-point–free symmetric path, matches NVFP4’s FP8 group scale + FP32 master scale, and fully uses all four codes at 2-bit.

\subsection{Muon for Low-bit QAT: Training Dynamics}
We investigated whether the \textit{Muon}~\citep{liu2025muon,park2025outliersafe} optimizer can stabilize training dynamics in ultra–low-bit QAT. In our pipeline, the per-group scale and zero-point are computed \emph{online}; thus the only trainable variables are the full-precision 2D weight matrices, while quantizer statistics are not explicitly optimized.

\textbf{Setup.} We keep the learning-rate schedule, batch size, and clipping identical to the AdamW baseline, and apply STE for quantization with progressive bit reduction.

\textbf{Observation.} Across models and bit settings, Muon did not yield consistent gains over AdamW: convergence speed and final perplexity were comparable or slightly worse, and we observed larger short-horizon oscillations near quantization thresholds in some layers.

\textbf{Possible causes (hypotheses).}
(i) Online rescaling induces non-stationary curvature that weakens Muon’s preconditioning benefits under STE noise; 
(ii) gradient signals are dominated by rounding discontinuities at ultra–low bits, reducing the utility of curvature-aware updates; 
(iii) block/group-wise statistic updates interact with momentum, amplifying drift.

\textbf{Next steps.} We will explore (a) using Muon only on LoRA adapters while freezing the backbone; (b) scale-aware trust-region or gradient clipping around threshold crossings; (c) layer-wise Muon/AdamW hybrids. At present, Muon does not provide a clear advantage for our low-bit QAT setting.

\section{Error Estimation for OCS}
\label{app:ocs_error}
To justify the efficacy of our Outlier Channel Split (OCS) strategy, we provide a formal error analysis comparing our \textbf{Rounding-aware (RA) split} with a \textbf{naive half split}. Consider a selected outlier channel $m$ with a weight row $W_{m:}$ and an input activation $x_m$. We decompose the weight into two branches with symmetric half-step offsets relative to the (post-split) step size $s$:
\begin{align*}
W_{m:}\ &\longrightarrow\ ( W_{m:}^{(1)}, W_{m:}^{(2)} ) \\
&= \Big( \frac{W_{m:} - s/2}{2},\ \frac{W_{m:} + s/2}{2} \Big).
\end{align*}
By nearest rounding, $Q_s(W_{m:}^{(1)}) + Q_s(W_{m:}^{(2)}) = Q_s(W_{m:})$. Defining the rounding error function as $\text{RoundErr}(z) = \text{Round}(z) - z \in [-\frac{1}{2}, \frac{1}{2})$, the resulting error $\varepsilon_{\text{RA}}$ can be derived as:
\begin{align*}
\varepsilon_{\text{RA}} &= x_m \left( Q_s(W_{m:}^{(1)}) + Q_s(W_{m:}^{(2)}) - W_{m:} \right) \nonumber \\
&= x_m \left( Q_s(W_{m:}) - W_{m:} \right) \nonumber \\
&= x_m \cdot s \cdot \text{RoundErr}\left( \frac{W_{m:}}{s} \right).
\end{align*}

In contrast, a \textbf{naive half split} $(W_{m:}/2, W_{m:}/2)$ forces each branch to be quantized independently without the benefit of offset cancellation. This results in a cumulative error $\varepsilon_{\text{naive}}$ governed by a coarser quantization scale $2s$:
\begin{align*}
\varepsilon_{\text{naive}} &= x_m \left( Q_s\left(\frac{W_{m:}}{2}\right) + Q_s\left(\frac{W_{m:}}{2}\right) - W_{m:} \right) \nonumber \\
&= x_m \left( 2 \cdot s \cdot \text{Round}\left(\frac{W_{m:}}{2s}\right) - W_{m:} \right) \nonumber \\
&= x_m \cdot 2s \cdot \text{RoundErr}\left( \frac{W_{m:}}{2s} \right)
\end{align*}
Hence $\mathbb{E}[|\varepsilon_{\text{RA}}|] = \frac{1}{2} \mathbb{E}[|\varepsilon_{\text{naive}}|]$, implying a 4$\times$ reduction in MSE. Assuming the splitting operation effectively halves the dynamic range such that the new step size $s \approx s_{\text{old}}/2$, our RA split achieves $\mathbb{E}[|\varepsilon_{\text{RA}}|] \approx \frac{1}{2} \mathbb{E}[|\varepsilon_{\text{base}}|]$, while the naïve split is even with the baseline.


\begin{table}[h]
\caption{Performance comparison on complex reasoning and instruction following capabilities (GSM8k, MathQA, MMLU, and IFEval)}
\label{tab:app_math}
\centering
\resizebox{0.48\textwidth}{!}{
\begin{tabular}{l c c c c c}
\toprule
& \textbf{Precision} & \textbf{Gsm8k} & \textbf{MathQA} & \textbf{Mmlu} & \textbf{Ifeval} \\
\midrule
Llama2-13B & FP16 & 0.22 & 0.32 & 0.52 & 0.18 \\
& \textsc{Bit-by-Bit} w2a16 & 0.20 & 0.32 & 0.50 & 0.17 \\
& \textsc{Bit-by-Bit} w2a2 & 0.11 & 0.29 & 0.40 & 0.16 \\
\midrule
Qwen2.5-7B & FP16 & 0.80 & 0.43 & 0.71 & 0.28 \\
& \textsc{Bit-by-Bit} w2a16 & 0.77 & 0.42 & 0.70 & 0.28 \\
& \textsc{Bit-by-Bit} w2a2 & 0.75 & 0.38 & 0.70 & 0.27 \\
\midrule
Qwen2.5-14B & FP16 & 0.84 & 0.52 & 0.77 & 0.32 \\
& \textsc{Bit-by-Bit} w2a16 & 0.84 & 0.51 & 0.75 & 0.30 \\
& \textsc{Bit-by-Bit} w2a2 & 0.81 & 0.50 & 0.75 & 0.30 \\
\bottomrule
\end{tabular}}
\end{table}

\section{Results on advanced reasoning and instruction following dataset}
Table~\ref{tab:app_math} presents the evaluation results on advanced reasoning and instruction-following benchmarks, including GSM8k, MathQA, MMLU, and IFEval. The Qwen2.5 family significantly outperforms Llama2-13B across all metrics, demonstrating superior mathematical reasoning and general knowledge capabilities. notably, Qwen2.5 models exhibit remarkable robustness to quantization. While Llama2-13B suffers a severe performance drop in the w2a2 setting (e.g., GSM8k score halving from 0.22 to 0.11), the Qwen2.5-14B maintains near-lossless performance, dropping only from 0.84 to 0.81. This indicates that the newer architecture is much more resilient to low-bit quantization in complex reasoning tasks.

\section{Ablation Study on W2A2 Setting}\label{app:w2a2_ablation}
\begin{table*}[ht]
    \caption{Ablation study on LLaMA-3.2-1B under the \textbf{W2A2} setting. All models are trained for one epoch. ``-'' indicates the component is disabled.}
    \label{tab:w2a2_ablation}
    \centering
    \resizebox{1.0\textwidth}{!}{
    \begin{tabular}{ccccccll}
        \toprule
        \rowcolor{gray!15}
        Block-wise & Progressive & Ocs & Metric & Calibration & group size & WikiText2 ppl & C4 ppl \\ 
        \midrule
        - & - & - & - & - & 32 & 2.0e5 & 1.0e6 \\
        \cmark & - & - & - & - & 32 & 1441.9 & 4592.8 \\
        \cmark & \cmark      & - & - & - & 32 & 42.2 & 120.2 \\
        \cmark & \cmark      & \cmark & Kurtosis & WikiText2 & 32 & 41.8 & 127.4 \\
        \cmark & \cmark      & \cmark & $w_{\max}$ & WikiText2 & 32 & 36.75 & 97.66 \\
        \cmark & \cmark      & \cmark & $\|\mathbf{x}\|_2 \cdot |\mathbf{w}|$ & WikiText2 & 32 & 32.48 & 79.95 \\
        \cmark & \cmark      & \cmark & $x_{\max}$ & WikiText2 & 32 & 32.34 & 76.79 \\
        \addlinespace[0.2em]
        \hdashline
        \addlinespace[0.2em]
        \cmark & \cmark      & \cmark & $x_{\max}$ & RedPajama & 32 & \textbf{31.82} & \textbf{72.63} \\
        \cmark & \cmark      & \cmark & $x_{\max}$ & C4 & 32 & 32.18 & 74.21 \\
        \cmark & \cmark      & \cmark & $x_{\max}$ & WikiText2 & 64 & 121.87 & 534.78 \\
        \cmark & \cmark      & \cmark & $x_{\max}$ & WikiText2 & 128 & 261.28 & 1191.11 \\
        
        \bottomrule
    \end{tabular}
    }
\end{table*}

To further validate the robustness and scalability of the \emph{Bit-by-Bit} framework under extreme quantization regimes, we provide a complementary ablation analysis under the \textbf{W2A2} (2-bit weight, 2-bit activation) setting. For these experiments, we train each configuration for only one epoch to facilitate rapid analysis. The results, including WikiText2 and C4 perplexity (PPL), are summarized in Table~\ref{tab:w2a2_ablation}.

\paragraph{Effectiveness of Progressive Strategy.} 
As shown in Table~\ref{tab:w2a2_ablation}, the baseline model without any of our proposed components fails to converge, resulting in a catastrophic perplexity (e.g., $2 \times 10^5$ on WikiText2). While the introduction of \emph{block-wise} optimization reduces the error, the perplexity remains unusable at $1441.9$. Crucially, the addition of our \textbf{progressive training} strategy brings the WikiText2 PPL down to $42.2$, representing a massive improvement in stability. This confirms that for ultra-low bit-widths like W2A2, the smooth optimization trajectory provided by the nested lattice structure is indispensable.

\paragraph{Comparison of Outlier Metrics.} 
We examine several metrics for identifying outlier channels for OCS. In the W2A2 regime, we observe that activation-based metrics are particularly effective. While the weight-only metric ($w_{\max}$) achieves $36.75$ PPL, the activation-centric metric $x_{\max}$ yields better robustness ($32.34$ PPL). This suggests that as activation precision drops to 2-bit, capturing and splitting activation outliers becomes more critical than in weight-only quantization. The joint metric $\|\mathbf{x}\|_2 \cdot \max|{w}|$ also performs competitively at $32.48$ PPL.

\paragraph{Impact of Calibration and Group Size.} 
Our analysis of calibration sets shows that using a sampled \texttt{RedPajama} subset yields the best alignment ($31.82$ PPL), likely due to its distribution being well-aligned with the data used during QAT. Regarding granularity, the W2A2 setting is highly sensitive to the group size. Increasing the group size from 32 to 128 leads to a sharp performance degradation, with WikiText2 PPL surging from $32.34$ to $261.28$. This underscores the necessity of fine-grained microscaling (e.g., group size 32) to maintain accuracy when both weights and activations are heavily quantized.

\section{Implementation details and More Results on GEMV}\label{app:operator}

This appendix provides low-level implementation details for our custom W2A2 GEMV kernel and the W2A16 kernel based on the Marlin framework, together with the exact test settings used in our evaluation.
We focus on the matrix-vector multiplication (GEMV) case during the decode stage, where a single activation vector multiplies a weight matrix:
\[
\mathbf{y} = \mathbf{x}\mathbf{W} \quad where \quad \mathbf{x}\in\mathbb{R}^{1\times K},\ \mathbf{W}\in\mathbb{R}^{K\times N}
\]

\subsection{GEMV kernel Implementation}
\subsubsection{W2A2 GEMV kernel}
\paragraph{Bit packing format and unpack with lop3.}
To maximize memory throughput, we store both weights and activations in a 2-bit packed format. Concretely, four 2-bit values are encoded into a single byte (\texttt{int8}).

We unpack packed 2-bit values into int8 lanes using a lightweight routine based on the lop3.b32 instruction, processing four elements per instruction. This significantly reduces the cost of unpacking in the W2A2 kernel.

\paragraph{Compute core: DP4A accumulation}
After unpacking, we compute dot products using integer SIMD instructions. Specifically, we use \texttt{dp4a} to accumulate products into an 32-bit accumulator before applying scaling and writing \texttt{bf16} outputs. This design keeps the compute pipeline lightweight while matching the packed 2-bit data layout.

\subsubsection{W2A16 GEMV kernel}
Our W2A16 kernel is built upon the Marlin framework, extending its optimized tiling and coalescing strategies to 2-bit regimes. The implementation stages activation and packed-weight tiles through shared memory with asynchronous copy, performs on-the-fly dequantization, and uses tensor-core MMA to accumulate in FP32 before applying per-column scales and writing FP16 outputs.

\begin{table}
\centering
\caption{Latency comparison for weight matrices appearing in the decode layers of LLaMA models.}
\label{tab:w2a2_bf16_llama_appendix}
\footnotesize
\setlength{\tabcolsep}{3pt}
\begin{tabular}{l c c c}
\toprule
Model & Shape $(N, K)$ & W2A2 ($\mu$s) & BF16 ($\mu$s) \\
\midrule
Llama3.2-3B
 & (1024, 3072)  & 8.93 &  6.45 \\
 & (3072, 3072)  & 8.64 & 16.04 \\
 & (3072, 8192)  & 8.70 & 32.80 \\
 & (8192, 3072)  & 8.94 & 20.76 \\
\midrule
Llama3-8B
 & (1024, 4096)  & 8.58 &  6.87 \\
 & (4096, 4096)  & 8.64 & 29.73 \\
 & (4096, 14336) & 11.56 & 133.49 \\
 & (14336, 4096) & 11.50 & 131.29 \\
\midrule
Llama3-70B
 & (1024, 8192)   & 9.34 &  8.57 \\
 & (8192, 8192)   & 9.02 & 150.15 \\
 & (8192, 28672)  & 18.62 & 509.98 \\
 & (28672, 8192)  & 19.39 & 516.95 \\
\bottomrule
\end{tabular}
\end{table}

\begin{table}
\centering
\caption{Throughput comparison on Llama3-8b}
\label{tab:w2a2_bf16_throughput}
\footnotesize
\setlength{\tabcolsep}{4pt}
\begin{tabular}{c c c}
\toprule
Sequence Length & BF16 (tokens/s) & W2A2 (tokens/s) \\
\midrule
64  & 49.02 & 76.59 \\
128 & 48.85 & 75.13 \\
256 & 48.59 & 74.97 \\
512 & 47.87 & 74.17 \\
\bottomrule
\end{tabular}
\end{table}

\subsection{Test setting and more results}
\label{app:gemv_test_more}
\paragraph{Test setting.}
Our GEMV kernel is fully written in CUDA 12.1 and compiled for NVIDIA Ada GPUs. All performance evaluations are conducted on a single NVIDIA RTX 4090 GPU. During our performance evaluations, we generate weights ($\mathbf{W}$) and activations ($\mathbf{A}$) corresponding to the designated low-bit precisions while keeping the effective compute shape identical across methods. Taking W2A2 as an example, we sample 2-bit weights and activations as integer tensors with values in $\{0,1,2,3\}$, and bit-pack them into a byte-packed 2-bit representation (i.e., four 2-bit values are stored in one byte) to form the packed activation/weight buffers before launching the custom kernel. For each configuration, we run the kernel 50 times and report the average latency. As a baseline, we time \texttt{bf16} GEMV using \texttt{torch.matmul} with matched shapes.

\paragraph{Results on more shapes.}
We benchmark a collection of \((N,K)\) shapes corresponding to projection layers in Llama3 family models. There are seven linear weight matrices:
$\mathbf{W}_{\text{down}}$, $\mathbf{W}_{\text{up}}$, $\mathbf{W}_{\text{gate}}$ in the MLP layer,
and $\mathbf{W}_{q}$, $\mathbf{W}_{k}$, $\mathbf{W}_{v}$, $\mathbf{W}_{o}$ in the self-attention layer.
Although there are seven distinct weights, they only instantiate four unique matrix shapes.
Specifically, $\mathbf{W}_{k}$ and $\mathbf{W}_{v}$ share the same shape,
$\mathbf{W}_{q}$ and $\mathbf{W}_{o}$ have identical shapes,
and $\mathbf{W}_{\text{up}}$ and $\mathbf{W}_{\text{gate}}$ also share the same dimensions,
resulting in four distinct shapes in total when including $\mathbf{W}_{\text{down}}$.

Table~\ref{tab:w2a2_bf16_llama_appendix} reports kernel latency for these four representative shapes across Llama 3.2-3B, Llama 3-8B and Llama3-70B models.
For matrices with relatively small output dimensions (e.g., $N=1024$), W2A2 exhibits slightly higher latency compared to the \texttt{bf16} baseline. This behavior is primarily due to fixed CUDA kernel launch overheads and the extra bit-level work required to unpack 2-bit operands, which dominate execution time in these regimes.
In contrast, for larger models and FFN-expanded shapes with large output dimensions, the workload exposes more parallelism, allowing W2A2 to better amortize launch overheads. The acceleration effect becomes significantly pronounced, exceeding $10\times$ speedup in some cases.
Furthermore, as table~\ref{tab:w2a2_bf16_throughput} shows, we evaluate the end-to-end decoding performance on the LLaMA3-8B model to demonstrate the practical efficacy of W2A2 in inference scenarios. Inference Speed (\textit{tokens/s}) is calculated as:
\[
\text{Speed} = \frac{\text{tokens}_{\text{prompt}} + \text{tokens}_{\text{generation}}}{\text{time}}
\]
We use a batch size of 1 and the minimum number of GPUs possible for evaluation. The speed is averaged over three runs.


\end{document}